# TARGET TRACKING IN THE RECOMMENDER SPACE
## *Toward a new recommender system based on Kalman filtering*

Samuel Nowakowski[1], Cédric Bernier[2]
[1]*LORIA – KIWI, Campus Scientifique - BP 239, 54506 Vandoeuvre Cedex, France*
[2]*Alcatel-Lucent Bell Labs France route de Villejust 91600 Nozay, France*
*Samuel.nowakowski@nancy-universite.fr, cedric.bernier@loria.fr*

Anne Boyer
*LORIA – KIWI, Campus Scientifique - BP 239, 54506 Vandoeuvre Cedex, France*
*Anne.boyer@loria.fr*

Abstract: In this paper, we propose a new approach for recommender systems based on target tracking by Kalman filtering. We assume that users and their seen resources are vectors in the multidimensional space of the categories of the resources. Knowing this space, we propose an algorithm based on a Kalman filter to track users and to predict the best prediction of their future position in the recommendation space.

## 1 INTRODUCTION

In Web-based services of dynamic content, recommender systems face the difficulty of identifying new pertinent items and providing pertinent and personalized recommendations for users.

Personalized recommendation has become a mandatory feature of Web sites to improve customer satisfaction and customer retention. Recommendation involves a process of gathering information about site visitors, managing the content assets, analyzing current and past user interactive behaviour, and, based on the analysis, delivering the right content to each visitor.

Recommendation methods can be distinguished into two main approaches: content based filtering (M Pazzani, D. Billsus, 2007) and collaborative filtering (D. Goldberg and al 1992). Collaborative filtering (CF) is one of the most successful and widely used technology to design recommender systems. CF analyzes users ratings to recognize similarities between users on the basis of their past ratings, and then generates new recommendations based on like-minded users' preferences. This approach suffers from several drawbacks, such as cold start, latency, sparsity (M. Grcar, D. and al 2006), even if it gives interesting results.

The main idea of this paper is to propose an alternative way for recommender systems. Our work is based on the following assumption: we consider Users and Web resources as a dynamic system described in a state space. This dynamic system can be modelled by techniques coming from control system methods. The obtained state space is defined by state variables that are related to the users. We consider that the states of the users (by states, we understand « what are the resources they want to see in the next step ») are measured by the grades given to one resource by the users.

In this paper, we are going to present the effectiveness of Kalman filtering based approach for recommendation. We will detail the backgrounds of this approach i.e. state space description and Kalman filter. Then, we expose the applied methodology. Our conclusion will give some guidelines for future works.

# 1. PRINCIPLES

In this part, we are going to describe the main principles of our approach, from the main hypothesis to the theoretical backgrounds.

## 1.1 Target Tracking in the Cyberspace

In this work, we assume that a user is a target moving in a specific space. The space will be defined by the main categories describing the viewed resources. This space called recommendation space will have as many dimensions as categories. Figure 1 shows what this trajectory looks like:

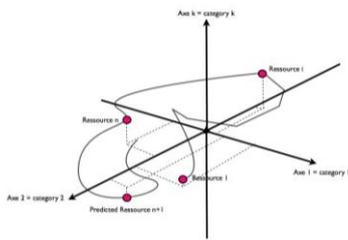

Figure 1. Space representation

Using this assumption, we are going to detail what is Kalman filtering, how we can apply it in this case and what are our main assumptions.

## 1.2 Kalman filter formulation

The basic task of the Kalman Filter is to estimate as accurately as possible the position and the velocity of a moving object. In our case, the moving object will be a user described in the new space of the categories of the resources.

First, we have to define the state vector of a user. Because our assumption (the user will a point in the space of the categories). This state vector at time k will be as follows:

$$X_k = \begin{vmatrix} x \\ \dot{x} \\ \ddot{x} \end{vmatrix}_k \quad (1)$$

Where

- $x$ is the vector containing the position coordinates
- $\dot{x}$ is the vector containing the velocity coordinates
- $\ddot{x}$ is the vector containing the acceleration coordinates

In a first step, we are going to use only the position vector x but the knowledge of $\dot{x}$ and $\ddot{x}$ will give us more details concerning the properties of the trajectories in the recommender space. Then, we can formulate the problem of target tracking by the following state space representation:

$$\begin{cases} X_{k+1} = AX_k + w_k & (2.a) \\ Z_k = HX_k + v_k & (2.b) \end{cases}$$

Where:

$$A = \begin{vmatrix} \alpha & T & \frac{1}{2}T^2 \\ 0 & \alpha & T \\ 0 & 0 & \alpha \end{vmatrix} \quad (3)$$

T can be seen as the mean time interval between two positions in the "cyberspace"

(Gibson, W.). T comes from the equation which links positions to speed and acceleration. Because it represents the time spent between to position (i.e. choices in the movies database), we put it equal to 1 (simulations have shown that its value does not influence the computations).

In the state space formulation given in equations (2.a) and (2.b), we have:

- $w_k$ is assumed to be Gaussian random vector which allows us to consider random behaviours of the observed users and $w_k \approx N(0,Q)$
- $v_k$ denotes perturbations on the measurements (in our case, this perturbations are minimized) $v_k \approx N(0,R)$

Measurement matrix H has the following structure:

$$44 \text{ rows} \begin{bmatrix} 1 & 0 & 00 & \ldots & 00 & \ldots & 0 \\ 0 & \ldots & 0\ldots & \ldots & 00 & \ldots & 0 \\ 0 & 0 & 10 & \ldots & 00 & \ldots & 0 \end{bmatrix}$$

(3 * 44 columns)

Knowing that we can derive the equations of the Kalman predictor. This predictor will be able to predict the future state on the trajectory. These equations are the following (for further details, concerning how to obtain these equations, see (Gevers, M. and Vandendorpe)):

The computed prediction is given by:

$$\begin{cases} \hat{X}_{k+1/k} = \hat{X}_{k/k-1} + K_k \left( Z_k - H\hat{X}_{k/k-1} \right) \\ = (A - K_k C)\hat{X}_{k/k-1} + K_k Z_k \end{cases} \quad (4)$$

The gain of the filter is:

$$K_k = AP_{k/k-1} H^T \left( HP_{k/k-1} H^T + R \right)^{-1} \quad (5)$$

The evolution of the uncertainty on the state estimation is given by:

$$P_{k+1/k} = AP_{k/k-1} A^T \quad (6)$$
$$- AP_{k/k-1} H^T \left( HP_{k/k-1} H^T + R \right)^{-1} HP_{k/k-1} A^T$$

Where:

- $\hat{X}_{0/-1} = X_0$ are the initial conditions
- $P_{0/-1} = P_0$
- $\hat{X}_{k+1/k}$ is the prediction of the state; it is the optimal estimation of the state of the model
- $\left( Z_k - H\hat{X}_{k/k-1} \right)$ is called the Innovation sequence
- $\hat{X}_{k+1/k}$ is the state prediction at time k+1 knowing states from time 0 to time k

Using this algorithm, we are going to consider our target as described in figure 2:

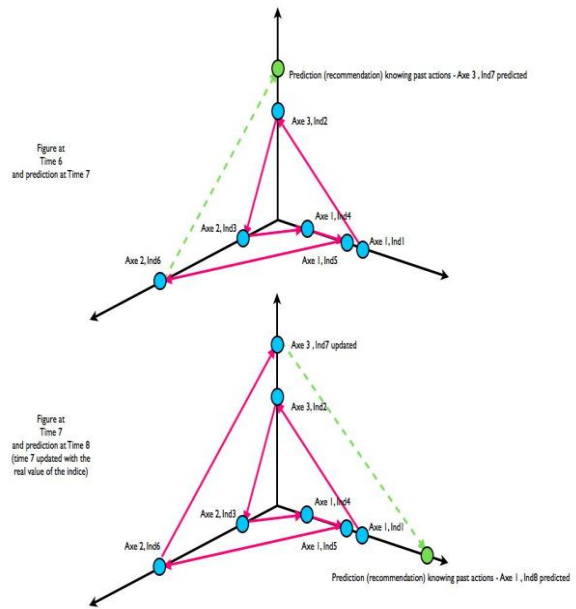

Figure 2. Trajectory in the recommendation space

Using Kalman filtering, we can obtain the best recommendation (predicted index of satisfaction) knowing all the past index seen as coordinates on the recommendation space.

## 2. APPLICATION

### 2.1 Description of the experiment

This experiment is based on TV consumption. The dataset is the TV consumption of 6423 english households over a period of 6 months (from 1st September 2008 to 1st March 2009) (Broadcaster Audience Research Board), (Senot, C. and al, 2010). This dataset contains information about the user, the household and about television program. Each TV program is labelled by one or several genres. In the experiment, a user profile build for each person. The user profile is the set of genres associated to the value of interest of the user for each genre. This user profile is elaborated in function of the quality of a user's TV consumption: if a TV program is watched entirely, the genre associated to this TV program increases in the user profile. Several logical rules are applied to estimate the interest of a user for a TV program.

The methodology of the experimentation is the following:

- Each user profile is computed at different instants (35) from the TV viewing data.
- The Kalman Filter is applied iteratively to estimate the following positions of the user profile in the space of the genre.

All the consumption is described by 44 types which will define the 44 dimensions space where users are "moving".

## 3. Numerical results

The obtained results can be exposed as follows:

- Kalman filter predicts the interest of a specific user for one gender knowing his past.

Using this prediction, we can propose a new recommendation strategy:

- If the Quantity of Interest (QoI) of the user is predicted to be in one specific region of the space, we can recommend something inside this specific region:
- For example, if the specific region is defined by dimensions Documentary and Drama, we can recommend contents related to these two dimensions
- If the predicted quantity of interest (QoI) changes to another dimension of the space, we can automatically recommend content from this new region of the space.

In the following figures (3 to 6), we can see the estimation/prediction given by the Kalman filter. The green lines show the prediction obtained at each time using the knowledge we have of the degree of interest of each user. Figure 7 shows the results of the cosine distance which has been computed between the true values and the prediction by the filter.

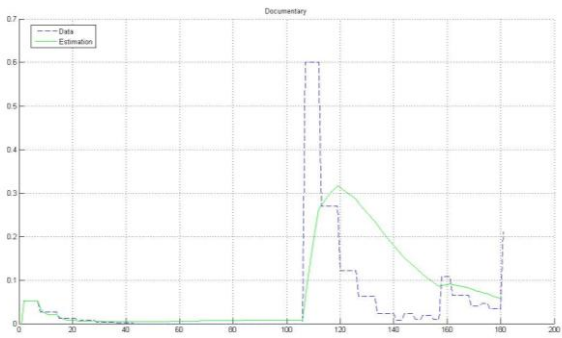

Figure 3. Prediction – index of interest – documentary

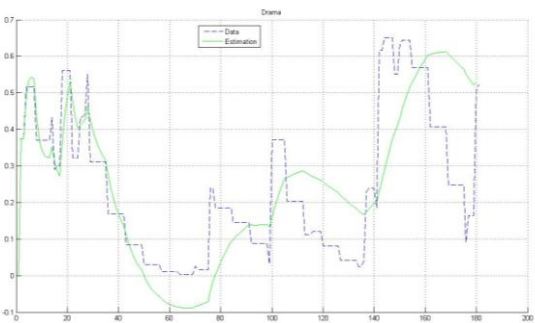

Figure 4. Prediction – index of interest – Drama

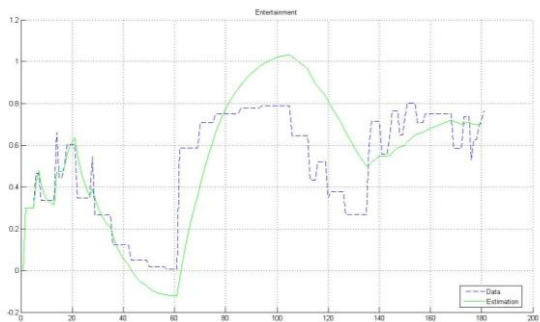

Figure 5. Prediction – index of interest - Entertainment

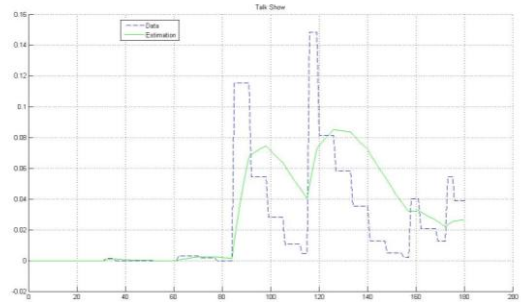

Figure 6. Prediction – index of interest – Talkshows

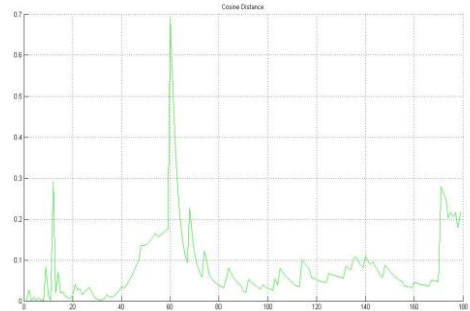

Figure 7. Distance between prediction and real measurements

Figures 3, 4, 5 and 6 show that the prediction follows the real measures of the index of interest related to all the observed categories. We note that it eliminates sudden changes and smoothes the abrupt variations. These characteristics are those we are going to use in our strategy which focuses on macroscopic recommendations (see next section).

The Figure 7 represents the cosine distance calculated between the estimation data and the observed data. Most of estimated values are corrects, with a cosine distance inferior to 0,15.

## 4. RECOMMANDATION PROCEDURE

In this approach, we can build a recommendation by analysing the estimation provided by Kalman Filter.

The profile is built from the consumption of TV programs. Each TV program is defined by concepts such as entertainment, science fiction, talk show, etc. The analysis of the way different TV programs are watched allows deducing the interest of a user for each concept. Hence, the user profile is calculated from the TV consumption and it is represented by a vector of valuated concepts.

The user profile is considered as a point in the space of the concepts (in our case, 44 dimensions). This point moves at each different time in the space and so forms a trajectory. With the Kalman Filter, we estimate the next position of the user profile.

The estimation shows the evolution within each dimension, hence for each concept.

For our new recommending strategy, we observe the difference between the estimated concept and the calculated concept. If the calculated concept is superior to the estimated concept (noted negative difference), then the user's interest for this concept is decreasing. On the contrary, if the estimated concept is superior to the calculated concept (noted positive difference), then the user's interest for this concept is increasing. We concentrate on the concepts showing up a big difference: the concepts with an important positive difference influence the recommendation towards these concepts, whereas the concepts with an important negative difference discourage the recommendation towards these concepts.

Conversely to existing methods which recommend precise contents for a given user, this method takes into account the user's state of mind. Our method performs on the macroscopic level. We find out the type of content the user appreciates and can determine some dimensions that can deliberately be closed out. The recommendation is based on the two preceding arguments:

- the user's actual state of mind
- the subset of retained dimensions.

Besides, we work on day-by-day data, hence we know the tendency of what the user would like to watch. We estimate the concepts the user will be interested in for the day. For example, the figure 8 shows that our system estimates that the user will be interested in the dimensions x y z but not at all in the dimensions alpha beta.

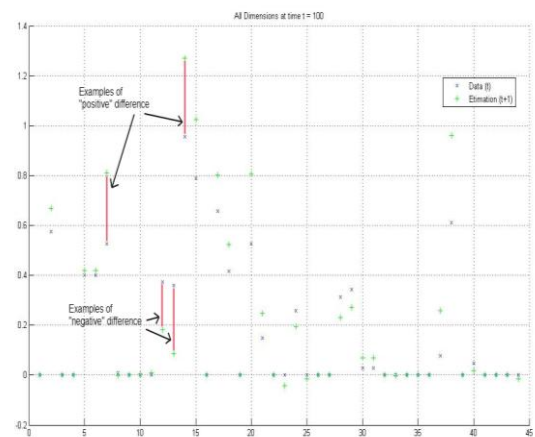

Figure 8. Analysis of the evolution of the prediction for recommendation

From these "positive" or "negative" dimensions and from the TV program, we have to define the recommendation for a set of TV programs for that day. Furthermore, according to what the user watched during the day, we can refine our recommendation. Indeed, in our example, if the user is interested in contents of types x, y and z and if he has already watched content of type x and y that day, the recommendation would essentially concentrate on content of type z.

Hence we need to make a last step which is to find a content which corresponds to the estimation of the dimensions' evolution.

## 4 CONCLUSIONS

In this paper, the main idea is to consider that the one who chooses films as a target which moves along a trajectory in the recommender space. The recommender space is seen as a 44 dimensions spaces based on the main concepts describing the films. The position of the target is measured by the index of interest expressed for each concept. Then the Kalman filter applied using a tracking model predicts the "positions" in the recommender space.

Then, knowing the past positions of the user in this space along the different axis of the 44 dimensions space, our Kalman based recommender system will suggest:

- if the user is interested in contents of types x, y and z and if he has already watched content of type x and y that day, the recommendation would essentially concentrate on content of type z
- knowing the position in the space, the best prediction for his next positions in the recommender space i.e. his best index of interest related to the favourite contents.

The strength of our approach is in its capability to make recommendations at a higher level which fit users habits i.e. given main directions to follow knowing the trajectory in the space and not to suggest specific resources.

Future works will be focused on tracking groups of users and on the definition of the topology of the recommendation space as a space including specific mathematical operators.